\documentclass[twoside]{article}

\usepackage[accepted]{aistats2021}
\usepackage{graphics} 
\usepackage{epsfig} 
\usepackage{mathptmx} 
\usepackage{times} 
\usepackage{amsmath} 
\usepackage{amssymb}  
\usepackage{subcaption}
\usepackage[normalem]{ulem}
\usepackage{todonotes}
\usepackage{hyperref}

\begin{document}


\runningauthor{ Qadeer Khan, Patrick Wenzel, Daniel Cremers }

\twocolumn[

\aistatstitle{Self-Supervised Steering Angle Prediction for Vehicle Control Using Visual Odometry}

\aistatsauthor{ Qadeer Khan\textsuperscript{1,2} \And Patrick Wenzel\textsuperscript{1,2} \And  Daniel Cremers\textsuperscript{1,2} }
\aistatsaddress{ \textsuperscript{1} Technical University of Munich    \\ \textsuperscript{2} Artisense \\ \texttt{\{qadeer.khan, patrick.wenzel, cremers\}@tum.de }}

]

\begin{abstract}
Vision-based learning methods for self-driving cars have primarily used supervised approaches that require a large number of labels for training. However, those labels are usually difficult and expensive to obtain. In this paper, we demonstrate how a model can be trained to control a vehicle's trajectory using camera poses estimated through visual odometry methods in an entirely self-supervised fashion. We propose a scalable framework that leverages trajectory information from several different runs using a camera setup placed at the front of a car. Experimental results on the CARLA simulator demonstrate that our proposed approach performs at par with the model trained with supervision.
\end{abstract}
\section{Introduction}\label{sec:introduction}

The success of current learning-based approaches on various tasks can be attributed to the availability of a large amount of labeled data. In most circumstances, data is being generated at a rate that is far superior than it is humanly possible to annotate it. Even if possible, it would be expensive, time-consuming, and prone to errors. Vision-based sensorimotor control tackles the following problem: Given an image captured by a camera, what should be the corresponding steering command? Controlling a car using supervised steering labels has shown fairly good performance~\cite{bojarski2016end,chen2015deepdriving,codevilla2018end,muller2006off}. However, this performance is mostly limited to the domain in which the data was collected~\cite{wenzel2018modular}. In order to generalize across a variety of conditions, a car fitted with drive-by-wire needs to be driven by an \emph{expert} driver for all such possible conditions. This is an in-feasible, inefficient, and expensive solution for the following reasons:

\begin{itemize}
    \item Dynamic conditions such as lighting and weather are beyond human control. Hence, data collection for a specific scenario cannot be pre-planned.
    \item The trained model may overfit to the specific position at which the camera is placed. Any deviation, in the position of the camera at test time, may lead to deteriorated performance. 
    \item Collection of supervised data requires drive by wire cars. Since many vehicles driven on the street by drivers do not have this option, we lose out on bootstrapping this huge treasure of valuable data. 
\end{itemize}

It may be argued that modern cars have the CAN bus which may be used to extract out important information including steering commands which can then be used for supervision. However, common people are not expected to modify their vehicle, if even allowed by the law of the country they reside in. Moreover, self-modifications also have implications on the warranty/insurance premiums. Therefore, we have referred to an \emph{expert} as a dedicated driver who has the necessary license and permits from the relevant traffic and regulatory authorities to drive a modified vehicle on public roads. Traditional people are not expected to hold these permits. 

In contrast to supervised approaches, an alternative would be to deploy reinforcement learning (RL) based techniques, wherein instead of collecting expert supervised labels, a reward function is defined~\cite{abbeel2007application,abbeel2004apprenticeship,lillicrap2015continuous}. The model learns a policy by a hit and trial method using the reward function as an indicator for updating the model's weights. However, this is an extremely data-inefficient solution as it requires far more sessions to train a model when compared with supervised techniques. Moreover, car manufacturers would be unwilling to use an unstable learning-based method for their autonomous functions. Recently, the authors of~\cite{kendall2019learning} demonstrated an example of using deep RL for training a policy for maintaining the car on the road in a controlled environment. However, this approach requires an expert driver to intervene for any correction whenever the car attempts to deviate away from the road. 

In this paper, we propose a self-supervised approach which that does not suffer from the limitations of the supervised and RL-based approaches as described earlier. We use a camera setup, that can be placed at the front of any car model. Ideally, we would like to retrieve the precise ground truth poses of the cameras as the car moves along its trajectory. This could be done using precise GNSS positions or through localization against landmarks. However, acquiring accurate camera poses using these methods may not always be possible such as when a car enters a tunnel or drives through an urban canyon due to the unavailability of GNSS signals or landmarks~\cite{weiss2011monocular}. Therefore, as an alternative, we demonstrate that using visual odometry (VO) methods~\cite{engel2017direct,mur2015orb,wang2017stereoDSO} to estimate the camera poses in a self-supervised manner can yield similar performance as with supervised labels. Note that, the VO methods can themselves be integrated with GNSS~\cite{inproceedingsgpsslam2009,Carlson2010MappingLU}, or inertial measurement units (IMUs)~\cite{articlevinsmono,stumberg18vidso} to improve the precision. However, to disambiguate the contribution of our method from additional sensor information, we only consider visual information, i.e. camera images.

\begin{figure}[t]
  \centering
  \includegraphics[width=\linewidth]{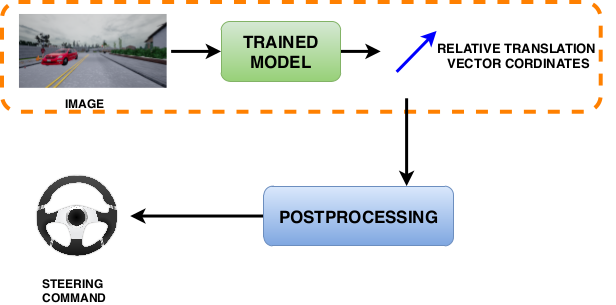}
  \caption{This figure shows the high-level overview for vehicle control. An image is fed to a deep neural network which is trained in a self-supervised manner using camera poses obtained by visual odometry. The model predicts the relative lateral translation vector for the next camera position, hence indicating the direction of where the car ought to move next. This vector value is then passed on to the post-processing module to produce an appropriate steering command.}
  \label{fig:highlevel_framework}
\end{figure}

The overall framework works as follows. Given an image, we train a model which predicts the lateral movement of the relative translation vector between 2 camera frames. The relative translation vector predicted by our model at inference time can be considered as a \emph{virtual} target indicating to the vehicle where to traverse next. Therefore, our framework can be seen as a waypoint-based control approach. However, the main difference to other approaches~\cite{Kaufmann2018DeepDR,Agarwal2019LearningTD} is that we do not require physical waypoints nor accurate GNSS reception at any given time. The authors of~\cite{bansal2019-lb-wayptnav} also use \emph{virtual} waypoints to navigate a robot indoors, however, they assume that the odometry is perfectly known. Figure~\ref{fig:highlevel_framework} shows the high-level framework of the proposed self-supervised vehicle control method. Our emphasis is on retrieving the relative translation vector which would be the same irrespective of which car the module is placed on. Hence, our approach of placing the camera module at the front of any vehicle can be scaled up to as many numbers of running cars on the road as possible without drive-by-wire limitations or regulatory restrictions. To this end, we make the following contributions:

\begin{enumerate}
    \item Ability to control a car in a self-supervised fashion without the need for an expert driver. This is achieved by using camera poses obtained through visual odometry methods.
    \item We show how a combination of trajectory information from multiple runs leads to improved performance.
    \item Using our proposed scalable solution, we demonstrate how our model can generalize even to variations in translation and rotation of the camera at inference time.
\end{enumerate}

Note that under dynamic weather and lighting conditions, VO methods may fail to furnish the correct camera poses. Therefore, we additionally show how our approach can be extended to such circumstances where VO may fail.

Although there are many publicly available data sets \cite{Geiger2012CVPR,  DBLP:journals/corr/CordtsORREBFRS16,RobotCarDatasetIJRR, waymo_open_dataset, DBLP:journals/corr/abs-1803-06184, wenzel2020fourseasons} focused on driving applications, they do not offer the possibility to test the online performance of self-driving cars necessary to evaluate our approach. As discussed in Section \ref{sec:experiments-results}, this is because the online performance of embodied agents does not necessarily correlate with the offline error metrics \cite{anderson2018evaluation, codevilla2018offline, ross2011reduction}. To circumvent this problem, we report results tested on the CARLA simulator \cite{DosovitskiyCoRL2017}. It offers the facility to conduct an online evaluation of our method. 

\section{Method}\label{sec:method}

In this section, we first describe the self-supervised control framework we use. Our approach focuses on low-level lateral control for vehicles. The problem we want to tackle is the following: given an image from a monocular camera at any instance in time, we would like to determine the immediate steering command where the car should maneuver next. This can be done by predicting a translation vector indicating the direction from where the car currently is to where it ought to be. But how do we get this translation vector without explicit supervision? 

\subsection{Camera Pose Estimation} \label{sec:cam_pose_estimation}
The only requirement for our approach is to have camera poses estimated by visual odometry (VO). This can be done by placing a camera setup at the front of a car recording the images as the driver controls the car. A VO method run on a sequence of these images yields the corresponding camera poses. 

The 6DoF pose can be estimated using a state-of-the-art VO method such as DSO~\cite{engel2017direct} or ORB-SLAM~\cite{orbslam2}. The camera poses are represented as transformation matrices $\mathbf{T}_i \in \mathbf{SE}(3)$, which are transforming a point from the world frame into the camera frame. To get the relative pose between two cameras at time $i$ and $i+1$, we can extract out the rotation matrix $\mathbf{R}$ and the translation vector $\mathbf{t}$ as follows:

\begin{align}
    \left[\begin{array}{rr}
    \mathbf{R} & \mathbf{t}  \\
    \mathbf{0} & 1 \\
    \end{array}\right] = \mathbf{T}_{i} \mathbf{T}_{i+1}^{-1},
\end{align}

where the rotation matrix $\mathbf{R}$ performs a rotation in Euclidean space which can be represented by yaw, pitch, and roll angles. The translation vector $\mathbf{t}$ denotes a translation around the $x, y, z$ axes. Since the camera is rigidly attached to the car, the relative pose between any 2 camera frames also gives the relative information between the car positions at these 2 instances. We further assume that the car has a planar motion with its state determined by $x, y$, and orientation angle in a global frame of reference. Throughout this paper, we will assume that we do not observe motion along the $z$ direction of the car, as well as a constant roll and pitch. 

The translation vector is defined as the relative vector between 2 camera poses. Note that in the global frame of reference, trajectories corresponding to the same steering commands may undesirably have different translation vectors. Therefore, to circumvent this issue, we redefine the frame of reference to a local coordinate system such that the forward motion of the car at any instance in time is always in the $x'$ direction, whereas any lateral motion is always in the $y'$ direction. The coordinates of the relative position vector in the local frame of reference are labeled as $dx$ and $dy$. Details of which can be found in the supplementary. Note that if the rate at which frames are received or the speed at which the ego-vehicle is moving is variable, then calculating the translation vector between 2 consecutive frames will yield inconsistent results. To overcome this, we select only those pair of frames that are a fixed distance $dx$ apart up to a certain tolerance. Hence, $dx$ can be considered to be a constant. 

\subsection{Bicycle Model for a Car}\label{sec:bicycle_model}
\begin{figure}[ht]
  \centering 
  \includegraphics[width=\linewidth]{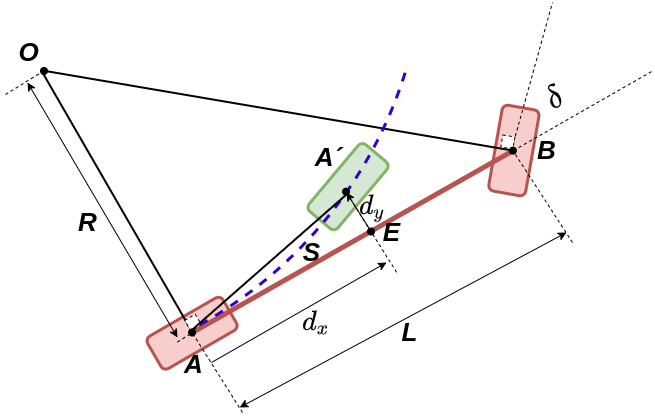}
  \caption{Depicts the bicycle model for a forward drive 4-wheeled car. $L$ is the distance between the rear (point \emph{A}) and front (point \emph{B}) wheels. With a steering angle of $\delta$, the car would traverse a curvature whose radius is given by $R$ and the rear wheel would move from point A to point A' by covering a distance $S$. The longitudinal and lateral motion of the car in the car's local frame of reference is given by $dx$ and $dy$ respectively. $O$ is the instantaneous center of rotation at the intersection of line segments drawn perpendicular to the front and rear wheels of the car. The dotted line represents the curvature of the trajectory.}
  \label{fig:bicyclemodel}
\end{figure}

We model the dynamics of a 4 wheel front-wheel drive car using the bicycle model \cite{wang2001} depicted in Figure \ref{fig:bicyclemodel}. The model assumes planar motion with the spatial coordinates and orientation describing the state of the car. The 2 rear wheels of the car are represented by a single rear wheel (point A). Likewise, the 2 front wheels are represented by a single front wheel (point B). The lateral motion of the bicycle can be controlled by maneuvering the steering angle ($\delta$) of the front wheel, whereas the longitudinal/forward motion of the vehicle is controlled by applying the throttle.

During turns, the front and rear wheels will be oriented in different directions. Attempting to make sharp turns at high speeds would exert excessive lateral forces, causing the car to slip. An important assumption we make is that the vehicle does not experience any slip. This is true when both wheels are oriented in the same direction irrespective of the speed. Whereas, while executing turns, this holds at moderately low speeds (5 m/s or $\approx$ 20 km/h) \cite{rajesh2012}. We can geometrically express the steering angle, $\delta$ in terms of the lateral translation vector determined by the method described in Section \ref{sec:cam_pose_estimation} as follows:

\begin{align}
    \angle OBA + 90 +  \delta &= 180 \\ 
    \angle OBA &= 90 -  \delta  \\ 
     \angle AOB + 90 +  \angle OBA &= 180 \\ \angle AOB &=  \delta  \\ 
     \tan(\angle AOB) &= \tan(\delta) = \frac{L}{R}
    \label{eq:model}
\end{align} 
\begin{equation}
    R = \frac{L}{\tan(\delta)}
    \label{eq:R_equation}  
\end{equation}

If the frame rate at which the information of successive frames is high coupled with the assumption of the vehicle moving at moderately low speeds while executing the turn would yield the following implications: 

\begin{enumerate}
\item The change in the orientation of the vehicle would be small giving $\Delta \phi$ $\approx$ $\angle$ A'AE $\approx$ 0.
\begin{align}
        \tan(\angle A'AE) \approx \tan(\Delta\phi) \approx \Delta\phi \approx \frac{dy}{dx}.
    \label{eq:dydx_equation}
\end{align} 

\item The distance $S$, traversed by the vehicle along the curvature is $\approx$ AA'. Moreover, $\Delta$$\phi$ $\approx$ 0 implies AA' $\approx$ $dx$.

\item The angular velocity of the car is given by \(\frac{d\phi}{dt}\) $\approx$ \(\frac{1}{R}\) $\cdot$ \(\frac{dS}{dt}\). A high frame rate implies: 

\begin{align}
    \Delta\phi \approx \frac{S}{R} \approx \frac{AA'}{R} \approx \frac{dx}{R}
    \label{eq:deltaphi_equation}
\end{align} 

Substituting Equation (\ref{eq:R_equation}) in (\ref{eq:deltaphi_equation}) results in

\begin{align}
    \Delta\phi = \frac{dx}{L} \tan(\delta)
    \label{eq:deltaphi_equation2}
\end{align} 

Equating Equations (\ref{eq:dydx_equation}) and (\ref{eq:deltaphi_equation2}) produces:

\begin{align}
    \frac{dy}{dx} = \frac{dx}{L}  \tan(\delta)\\
    \delta = \tan^{-1}(\frac{dy \cdot L}{dx \cdot dx})
    \label{eq:dx}
\end{align} 
\end{enumerate}

As described in Section \ref{sec:cam_pose_estimation}, the image pairs are selected such that they are approximately a distance of $dx$ apart. Hence, denominator of Equation (\ref{eq:dx}) can be taken as constant. Since $L$ is the length between the front and rear wheels, it is fixed, hence the above equation can be rewritten as:
\begin{align}
    \delta = \tan^{-1}(dy \cdot \alpha)
    \label{eq:final}
\end{align} 

Where $\alpha$ is a constant that can be tuned at run-time depending on the car. Therefore, we can determine the steering angle of the vehicle from the lateral component of the relative translation vector which in turn is determined from the method described in Section \ref{sec:cam_pose_estimation}. For this, we train a neural network model which takes in an image as input and outputs the corresponding value of $dy$ by minimizing the $L1$ loss between the value predicted by the model and value of $dy$. In the supplementary material, we demonstrate the implications on the performance of the trained models upon deviating from the assumptions given in this section.

\subsection{Multi-Car Trajectories}
\begin{figure}[ht]
  \centering
  \includegraphics[width=\linewidth]{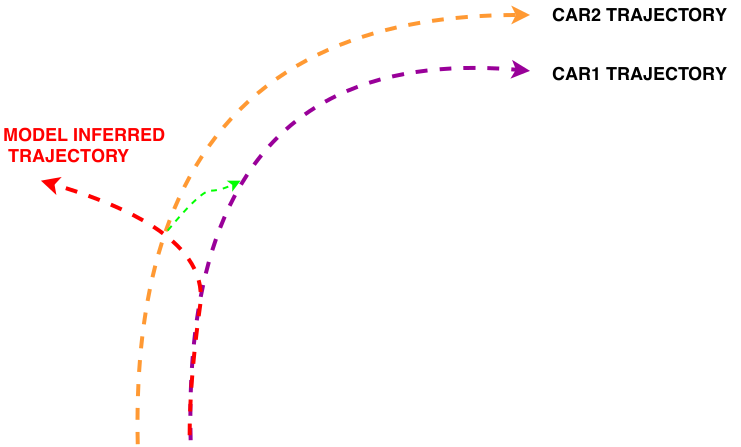}
  \caption{The purple line shows a section of the trajectory for \emph{Car1} used to train the self-supervised model. At test time, the model might slightly deviate away from this learned trajectory and come across images it has never seen in the training set. Hence, it would not be able to take the correct action to return to the original path after this slight deviation. Eventually, this deviation would accumulate with the car possibly crashing into an obstacle. The orange line shows the trajectory of another different car (\emph{Car2}). During training,  images from this trajectory can also be used to map the relative vector from \emph{Car2} trajectory to the next frame of \emph{Car1}. Hence, at test time, even if the car deviates from the original \emph{Car1} trajectory, it will still be able to make the correct maneuver to return to its original course.}
  \label{fig:deviated_trajectory}
\end{figure}

One issue with training a self-supervised vehicle control model with only a single car trajectory is that there may not be many variations in the collected data. Hence, at inference time, the car may slightly deviate from the reference trajectory (being the only trajectory on which it was trained) and come across a scene of possibly a different trajectory which it had never seen in the training set. Hence, the model may not be able to figure out the correct action to take against the next sequence of images. This is depicted in Figure \ref{fig:deviated_trajectory}.

To solve for this, note that drivers have a unique driving preference. Some tend to drive in the center, while others close to the lane or sidewalks. In fact, the same driver might demonstrate different trajectories while driving along the same route multiple times. Note that these behaviors lead to a diverse set of trajectories along the same section of a road. Hence, a slight deviation for one driver may be the usual trajectory for a different driver. But how can we scale our training to also incorporate this vast amount of information being generated by multiple drivers/runs across the same road section? This is also depicted in Figure~\ref{fig:deviated_trajectory} by the orange curve showing the trajectory generated by a different run along the same route. For any image along this new trajectory, the model aims to predict the relative position vector to the next closest image pose of the original purple trajectory. The frames in this new trajectory can be relocalized against the reference using direct visual localization approaches such as GN-Net~\cite{gn-net-19} or by running the PnP algorithm in a RANSAC scheme using robust visual image descriptors, e.g. R2D2~\cite{r2d2}.

\noindent{\emph{Reference Trajectory:}} For any section of the road having multiple trajectories, we define the reference trajectory as the one to which the relative vector from any of the many trajectories is mapped to. In Figure~\ref{fig:deviated_trajectory}, the purple line is taken as the reference trajectory. For the case of multiple trajectories, then for each, we calculate the lateral motion vector to every other trajectory and sum it up. The centermost will have the lowest sum and is taken as the reference. 

\subsection{Control Under Dynamic Weather/Lighting Conditions}\label{sec:teacher_student}
\begin{figure}[t]
  \centering
  \begin{subfigure}{0.49\linewidth}
    \includegraphics[width=\linewidth]{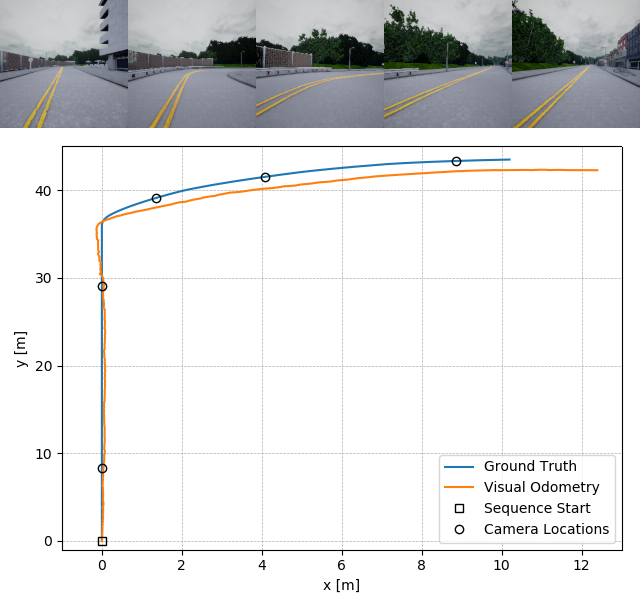}
    \caption{\emph{CloudyNoon} Trajectory.}
  \end{subfigure}
  \hfill
  \begin{subfigure}{0.49\linewidth}
    \includegraphics[width=\linewidth]{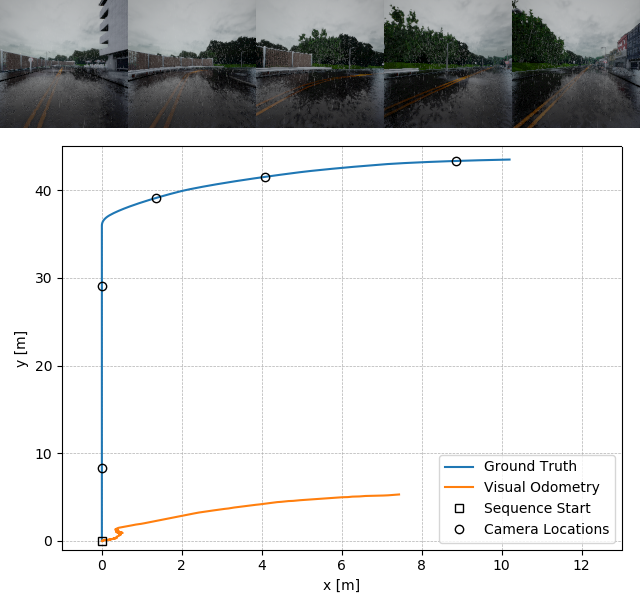}
    \caption{\emph{HardRainNoon} Trajectory.}
  \end{subfigure}
  \caption{This figure shows the influence of inclement weather on a small sample stretch of a trajectory produced by VO on the CARLA simulator. The curve in blue is the ground truth trajectory, and the curve in orange is the one obtained with VO. VO fails on the \emph{HardRainNoon} condition. Top: Camera snapshots from the respective locations on the sequence indicated by circles in the ground truth trajectory.}
  \label{fig:diff_weathers}
\end{figure}

In dynamic weather conditions, VO methods may fail to generate an accurate representation of the trajectory. Hence, it may not be possible to obtain self-supervised training data for controlling a car in such a scenario. This is depicted in Figure~\ref{fig:diff_weathers}, where under the \emph{CloudyNoon} condition the trajectory generated by VO matches closely with the ground truth trajectory. This is in contrast to the \emph{HardRainNoon} condition, where visual odometry fails for the same sequence. 

\begin{figure}[t]
  \centering
  \includegraphics[width=\linewidth]{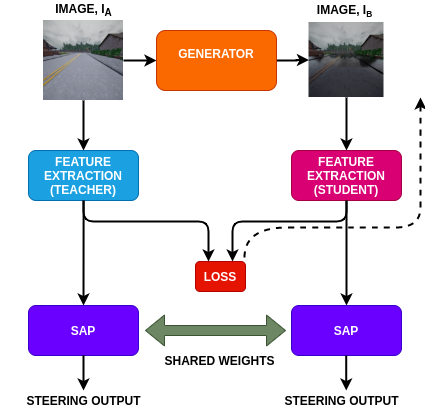}
  \caption{This figure shows the teacher-student architecture used to produce the steering commands independently of the image input domain. The generator is pre-trained using samples from both domains. The solid arrows represent forward propagation into the network, whereas the dotted line depicts back-propagation to update the weights of the student model.}
  \label{fig:teacher_student}
\end{figure}

To overcome this issue, we adopt the teacher-student approach proposed by~\cite{wenzel2018modular}. Note that our model can be divided into a feature extraction (FEM) and Steering Angle Prediction (SAP) module. The FEM extracts out features with an RGB image as input. These features are in turn fed to the SAP which predicts the coordinate for the lateral motion of the car. This model is trained with the visual odometry trajectories obtained under the \emph{CloudyNoon} condition and is referred to as the teacher. This model will fail if tested on \emph{HardRainNoon} condition. For this, we train another student model whose SAP module has exactly the same weights as that of the teacher model. Meanwhile, its FEM module is trained by feeding an image translated from \emph{CloudyNoon} to \emph{HardRainNoon} using an image-to-image translation (GAN) network~\cite{anoosheh2018night} trained in an unsupervised manner. The generator of the GAN takes in an image from domain $A$, $I_A$ (\emph{CloudyNoon}) and translates it to an image $I_B$ belonging to domain $B$ (\emph{HardRainNoon}). The weights are updated by minimizing the mean square error between the features of the 2 models. The architecture is depicted in Figure~\ref{fig:teacher_student}.

\section{Experiments}\label{sec:experiments-results}

Coming up with good metrics for evaluating the task of visual navigation is rather challenging. In particular, for autonomous driving, it is not obvious how to measure the performance of the driving policy. The authors of~\cite{anderson2018evaluation} present different problem scenarios for the analysis of embodied agent navigation and suggest some evaluation metrics for standard scenarios. Moreover,~\cite{codevilla2018offline} conduct further studies on online and offline metrics, especially for vision-based sensorimotor control. Specifically, in this work, the difference between metrics on static images compared to evaluating online in the simulation is analyzed. Their result showed that driving models can have similar offline metrics but drastically different driving performance. They concluded from their experiments that offline metrics are weakly correlated with actual driving quality. Therefore, it is not trivial to link from offline to online performance due to a low correlation between them. 

Due to the aforementioned reasons, we interpret our conclusions in an online setting as this is directly correlated to the actual driving performance. For this, we use the stable version v0.8.2 of CARLA~\cite{DosovitskiyCoRL2017} simulator for training and evaluation of our models. CARLA provides 15 different weather conditions. We train our models only on the \emph{CloudyNoon} condition but test our results on all other conditions. We compare the performance of our method with that of \cite{control-across-weathers-19} which trains a supervised model that is robust across all the weather conditions. The results were reported for executing turns around corners at fixed speeds. The metric to measure the performance of the models was done by calculating the ratio of time the car remains within its own driving lane. We adapt the same for comparison in our experiments.

\subsection{Models}
For the purpose of evaluation, we evaluate 4 different models whose descriptions are given below. While in principle, our approach can be deployed in conjunction with any VO based method, we have used \cite{wang2017stereoDSO} for our experiments:

\begin{figure}[t]
  \centering
  \includegraphics[width=\linewidth]{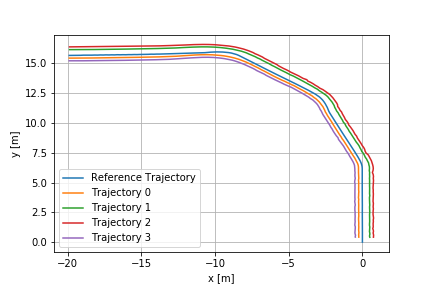}
  \caption{This figure shows the reference trajectory obtained by running the VO algorithm. 4 additional trajectories in relation to the reference are also depicted. }
  \label{fig:relo_multitrajectories}
\end{figure}

\begin{enumerate}
    \item \textit{Supervised model with ground truth steering angle:} This is the method of \cite{control-across-weathers-19} trained in a supervised manner using the ground truth steering angles directly obtained from the CARLA simulator. The network is capable of handling images from all 15 weather conditions.  We may use this model trained with ground truth supervised steering angle labels as the Oracle to compare against the other models not trained with any ground truth data. 
    \item \textit{Visual Odometry, with Multiple Trajectories:} Here we use the estimated camera poses obtained through VO to implicitly map to the steering angle of the car. Figure~\ref{fig:relo_multitrajectories} shows the reference VO trajectory along with 4 other trajectories in relation to the reference.
    
    \item \textit{Visual Odometry, with Single Trajectory:} Similar to the previous model. The images used for training this model are only taken from the reference trajectory.
    \item \textit{Visual Odometry, with Multiple Trajectories, Robust:} Same as the model described previously using VO on multiple trajectories except this model is also made robust to all the remaining weather conditions in a completely unsupervised manner by the method described in Section \ref{sec:teacher_student}. However, to cater for multiple weathers not only the generator for conversion between \emph{CloudyNoon} to \emph{HardRainNoon} is trained but also the generators for \emph{HardRainSunset}, \emph{WetCloudyNoon}, \emph{CloudySunset}, \emph{MidRainSunset}, \emph{ClearSunset}, \emph{ClearNoon}, \emph{WetSunset}, \emph{WetCloudySunet}, \emph{WetNoon}. Images produced by all these generators are used for training the student model.
\end{enumerate}

\begin{figure}[t]
  \centering
  \includegraphics[width=\linewidth]{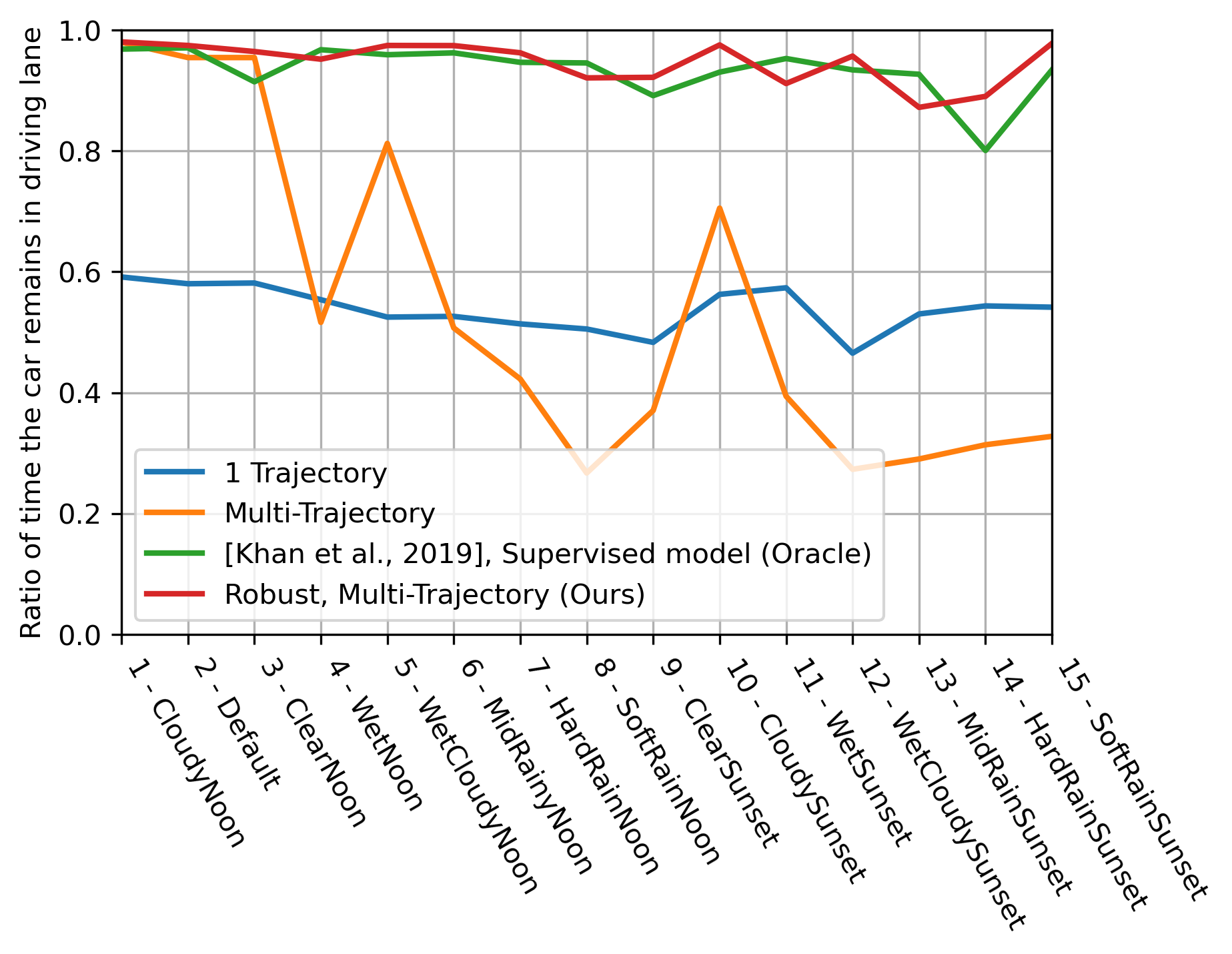}
  \caption{The plot exhibits the ratio of time the car remains within its own driving track for the 4 models across all the 15 different weather conditions. Higher is better.}
  \label{fig:all_weather_results}
\end{figure}

Figure \ref{fig:all_weather_results} is a plot showing the performance of the aforementioned 4 models against all the 15 weather conditions.

\section{Discussion}\label{sec:discussion}

In the following, we will elaborate on important key points regarding the performance of the 4 models and conduct additional studies to further investigate the findings. 

\begin{enumerate}
    \item \noindent{\bf Single Trajectory Performance:} All models demonstrate good performance on the weather condition on which they were trained with the model trained on 1 trajectory being an exception. This is because, during the online evaluation, slight deviations from the reference trajectory might lead to a collapse of the model since it has no way of coming back to its original course. This is due to the fact that the model had never seen such scenes from different trajectories during training. Another reason that would explain this poor performance is that if the poses estimated by visual odometry used to train this model are by themselves not perfect, then the model is being trained to the imperfections of this imprecise dataset. Although these predictions may be marginally incorrect locally, however, in an online setting they tend to accumulate over the long run and thereby causing a collision with obstacles. 
    
    \item \noindent{\bf Multi Trajectory Performance:} In contrast, the 2 models trained with VO on multiple trajectories exhibit superior online performance alluding to the utility of combining information from multiple trajectories. In fact, the performance of these 2 models is comparable with the Oracle that was trained in a supervised manner using ground truth labels.
    
    \item \noindent{\bf Method Scalability:} A plausible explanation for this is depicted in Figure \ref{fig:camseval} which demonstrates the scaling capability of our method. The blue curve shows that increasing the number of training trajectories improves the online performance. In fact, after a certain number of trajectories, the performance tends to saturate since any addition is already captured by the previous. Moreover, the figure also shows that our method is robust to changes in camera parameters at inference time. Hence the position/orientation of the camera can be adjusted to yield the best performance at inference.

    \item \noindent{\bf Performance under Dynamic Weather Conditions:} Note that both the models trained on multiple trajectories give good performance on the weathers that closely resemble the \emph{CloudyNoon} condition. However, on all other conditions, only our model trained for domain adaptation in a completely unsupervised manner by the method described in Section \ref{sec:teacher_student} maintains equivalent performance. It is interesting to note that our approach even works for the \emph{HardRainNoon} condition, where, VO had failed to produce the correct trajectory of the car as depicted in Figure \ref{fig:diff_weathers}.

\begin{figure}[t]
  \centering
  \includegraphics[width=\linewidth]{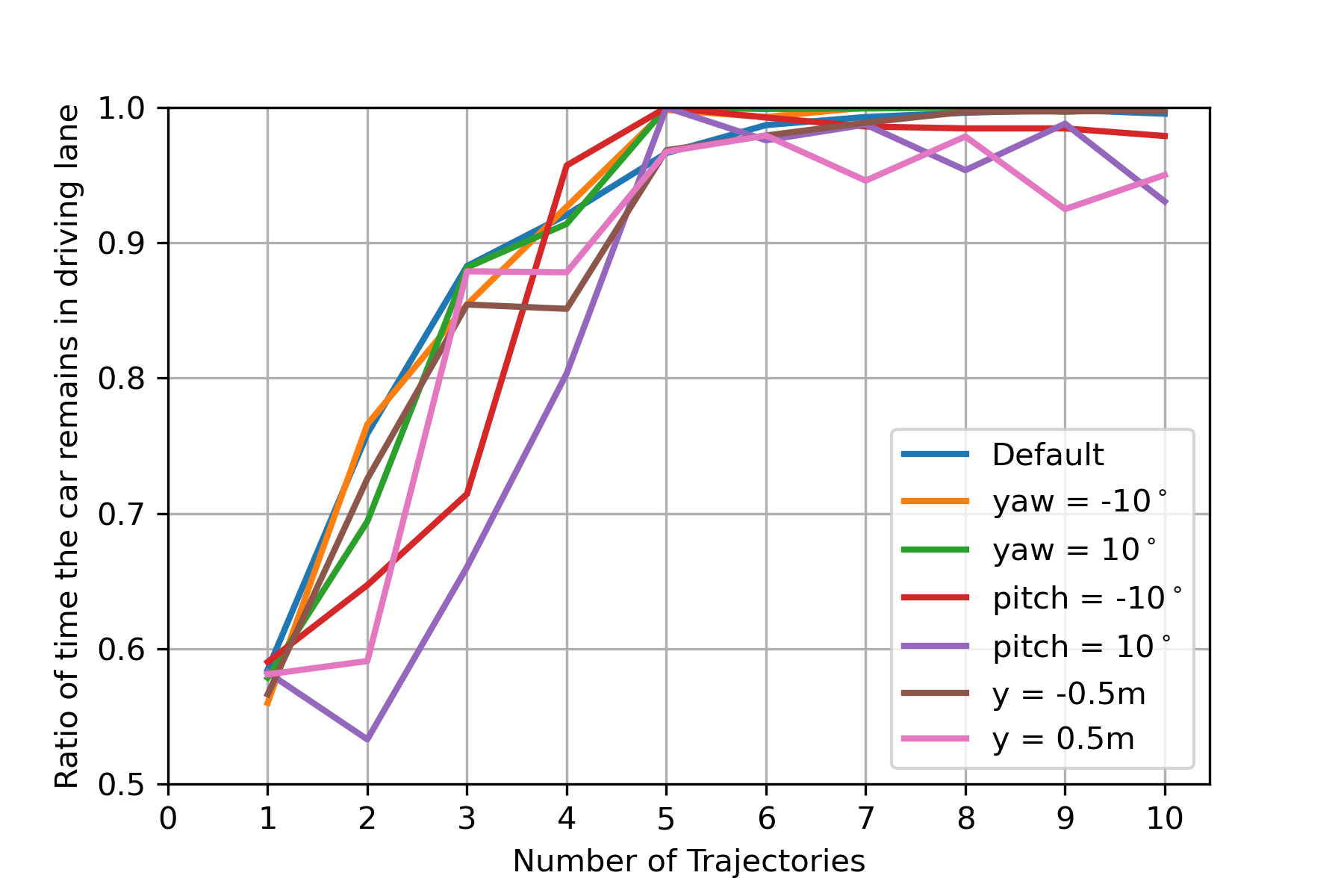}
  \caption{This figure shows the ratio of the time the car remains within its own driving track for different number of trajectories and changes in camera parameters from the default. The yaw and pitch values report the degree change in these parameters from the default. Whereas the relative movement of the camera along the width of the car ($y$) with respect to the default position is given in meters. Higher is better.}
  \label{fig:camseval}
\end{figure}

    \item \noindent{\bf Robustness to Perturbations:} Figure \ref{fig:jitter_all} shows the implication of inducing varying degrees of perturbations to the final steering command. As the perturbations are progressively increased, the performance for the model trained with only one trajectory, tends to correspondingly deteriorate. The models trained with multiple trajectories are less influenced by these perturbations. This can be observed in Figure \ref{fig:jitter_mean_std} wherein, not only the mean accuracy tends to improve but the deviation in performance across the perturbations also shows a diminishing trend with the increasing number of trajectories.
    
\begin{figure}[t]
  \centering
  \includegraphics[width=\linewidth]{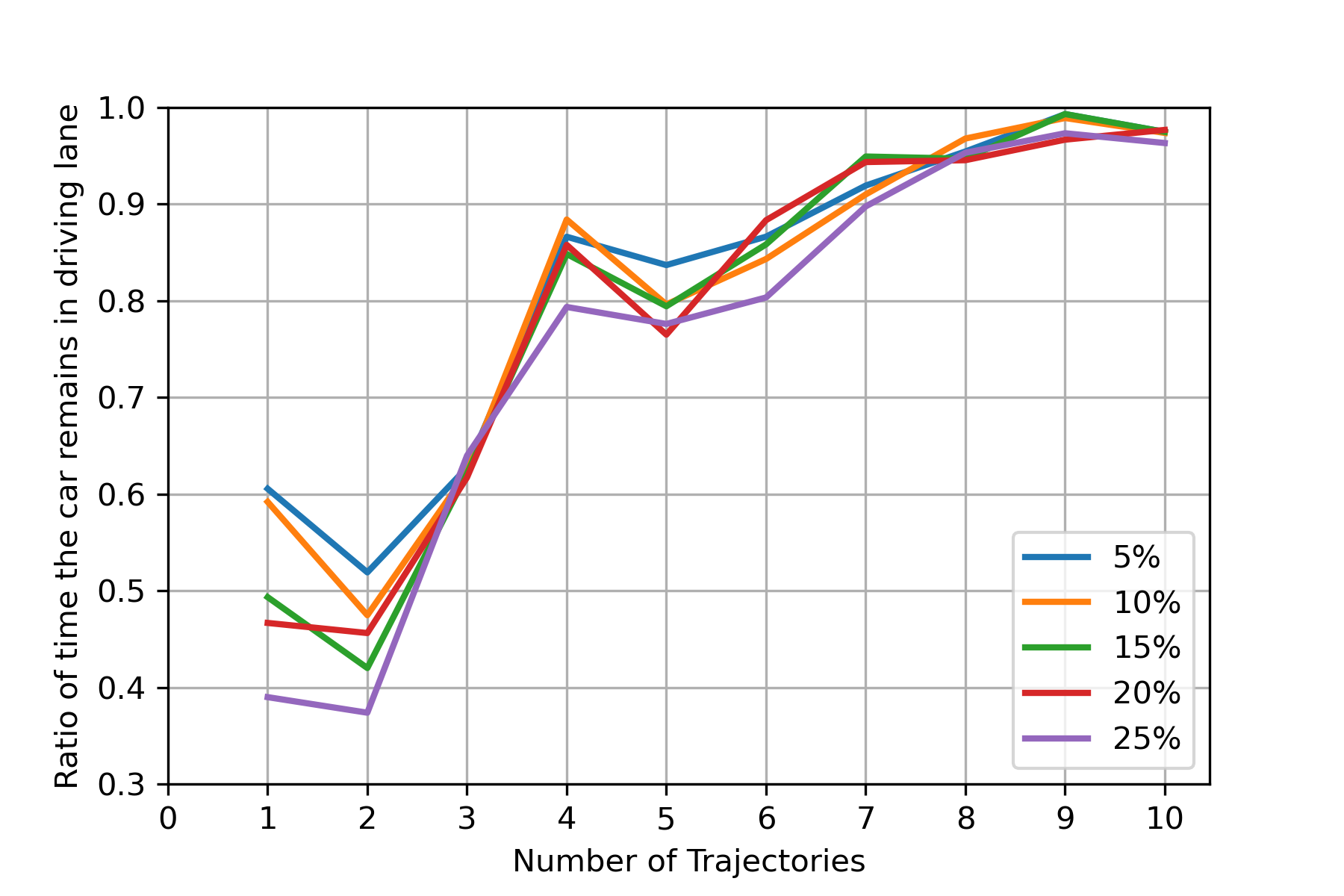}
  \caption{This figure shows the influence on the performance of the models trained with different number of trajectories as varying levels of perturbations (shown as a percentage) are introduced into the final steering command. For a specific percentage of perturbation, each line in the figure represents the mean ratio of the time the car remains within its own driving track across all the camera positions described in Figure \ref{fig:camseval}.}
  \label{fig:jitter_all}
\end{figure}
    
    \item \noindent{\bf Influence of Scalability on Perturbations:} It is interesting to note from Figure \ref{fig:camseval} that models which were trained with 5 or more trajectories maintain a performance ratio of remaining in the driving lane of greater than 0.9. However, the model trained with 5 trajectories is inferior in performance when perturbations are introduced as depicted in Figure \ref{fig:jitter_mean_std}. Here, a minimum of 8 trajectories was needed to achieve similar performance. This alludes that scaling our method across more trajectories tends to lead to better robustness to perturbations.
    
    \begin{figure}[t]
      \centering
      \includegraphics[width=\linewidth]{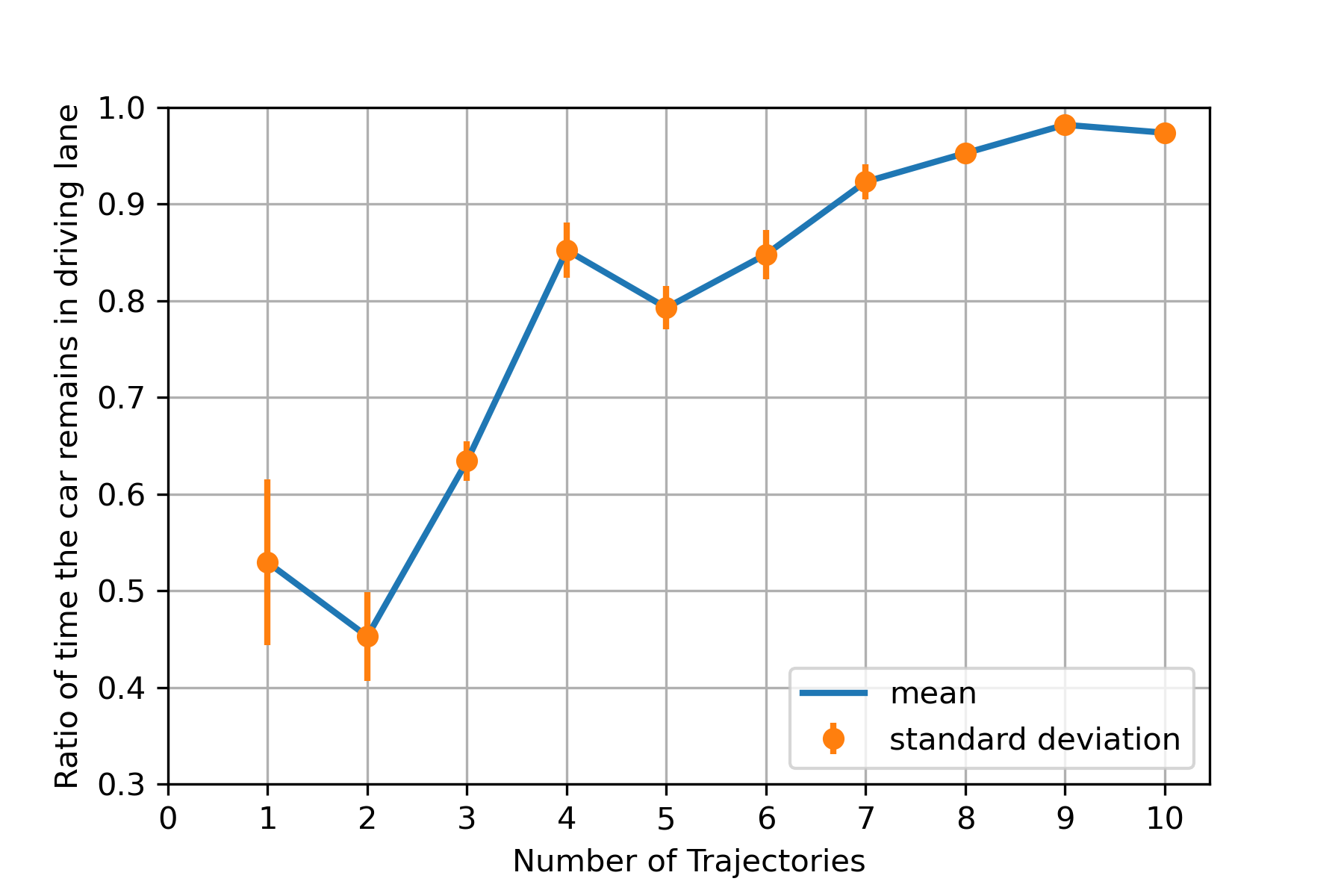}
      \caption{This figure shows the mean and standard deviation for the curves shown in Figure \ref{fig:jitter_all} across the different levels of perturbations.}
      \label{fig:jitter_mean_std}
    \end{figure}
\end{enumerate}

The online performance of the models along with a short summary of the proposed method can be 
found at this link: \url{https://youtu.be/_22gF9GuwTg}.
\section{Conclusion}\label{sec:conclusion}

In this work, we address a central challenge in the development of self-driving cars -- namely the question of how a machine can learn suitable driving strategies merely by observing the driver's behavior. We presented a novel approach for self-supervised vehicle control that combines the strength of any state-of-the-art visual odometry method with learning-based techniques. We achieve this in a framework that uses camera poses traced out along a vehicle's trajectory. Hence training can be done in an entirely self-supervised fashion. Moreover, our method of using multiple trajectories allowed for the scaling of the model. The proposed framework facilitates replacing the time-consuming and expensive process of collecting expert-driving data. This approach could be particularly useful for deploying on public transportation vehicles that have different drivers traversing the same route multiple times.

\bibliography{main}

\end{document}


\runningtitle{Self-Supervised Steering Angle Prediction for Vehicle Control Using Visual Odometry}

\runningauthor{ Qadeer Khan, Patrick Wenzel, Daniel Cremers }

\twocolumn[

\aistatstitle{Supplementary Material:\\Self-Supervised Steering Angle Prediction\\for Vehicle Control Using Visual Odometry 
}

\aistatsauthor{ Qadeer Khan\textsuperscript{1,2} \And Patrick Wenzel\textsuperscript{1,2} \And  Daniel Cremers\textsuperscript{1,2} }
\aistatsaddress{ \textsuperscript{1} Technical University of Munich    \\ \textsuperscript{2} Artisense \\ \texttt{\{qadeer.khan, patrick.wenzel, cremers\}@tum.de }}

]


\section{Global to Local Frame Transformations} 

\begin{figure}[ht]
  \centering
  \includegraphics[width=\linewidth]{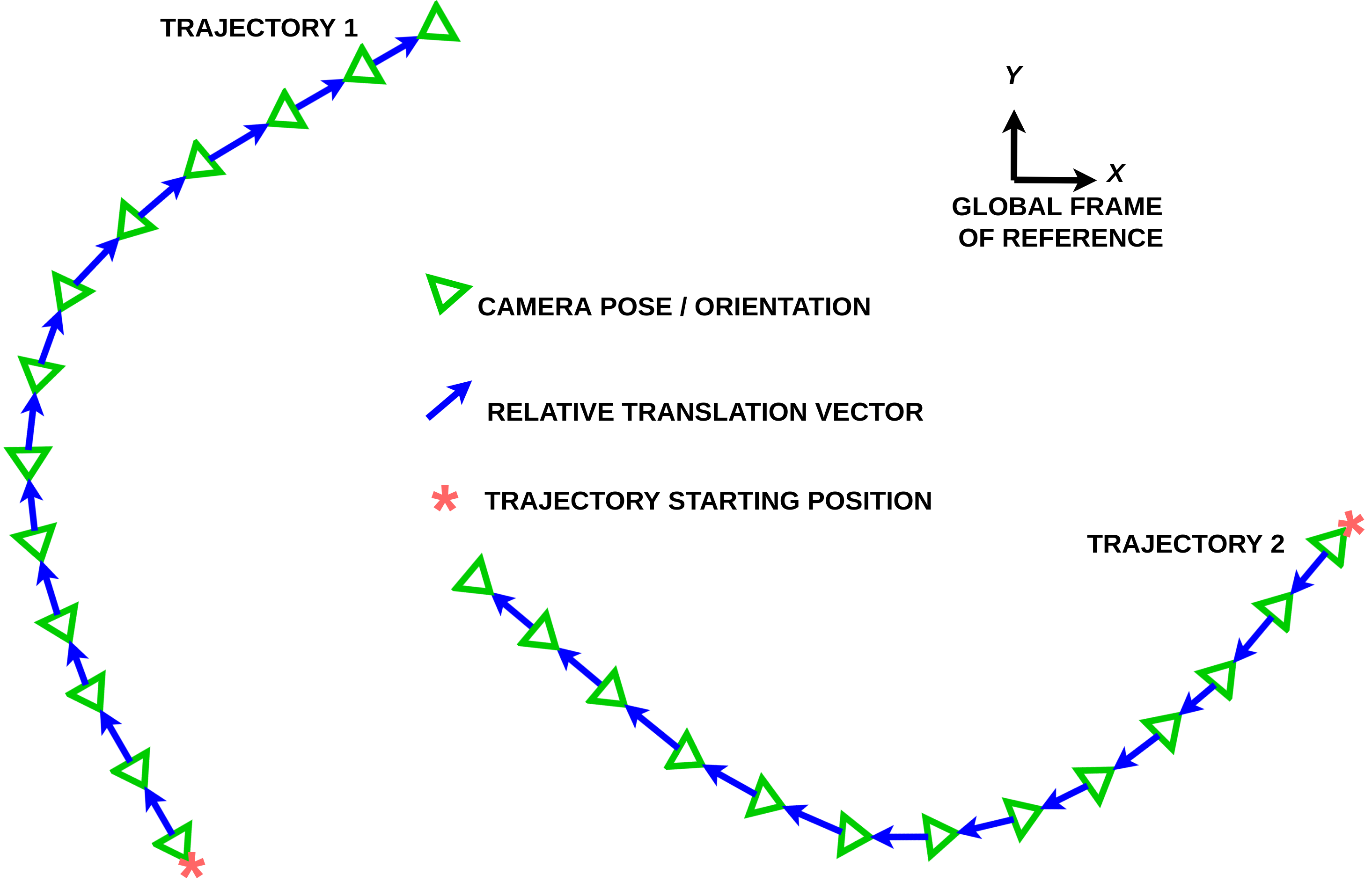}
  \caption{This figure shows two trajectories traversed by the ego-vehicle in the global frame of reference. Despite executing the same sequence of steering commands, the corresponding relative translation vectors point in different directions.}
  \label{fig:global_refProb_DriftNoprob_combined}
\end{figure}

Figure~\ref{fig:global_refProb_DriftNoprob_combined}, shows two trajectories generated using the same steering commands from two different starting positions. Note that in the global frame of reference, the two paths despite having the same steering commands produce different relative translation vectors. This is due to the fact that the 2 trajectories are oriented differently in the global frame of reference. 

To deal with this, we redefine the vectors to a local frame of the reference such that the forward direction is always in the $x'$-direction and the lateral movement is defined in the $y'$-direction. 

\begin{figure}[t]
  \centering
  \includegraphics[width=\linewidth]{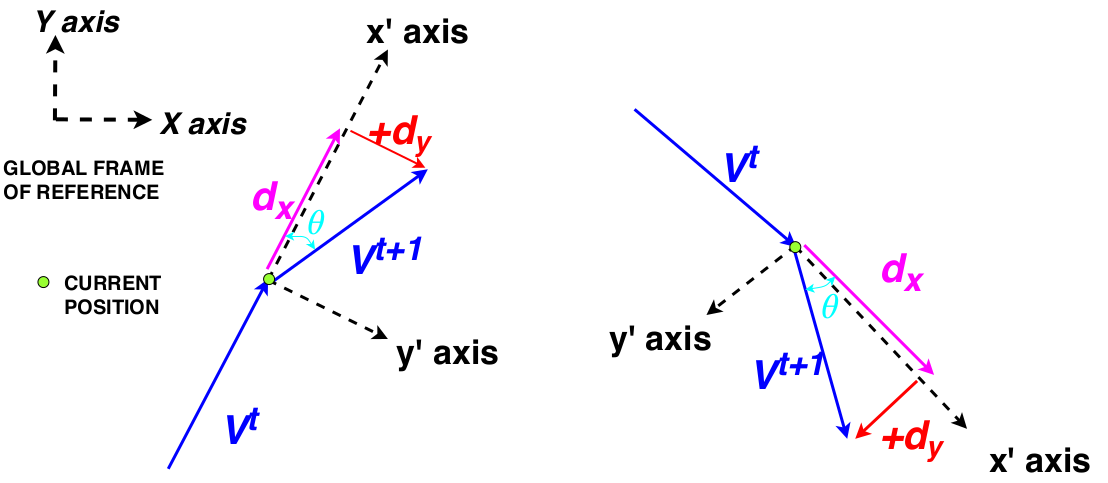}
  \caption{This figure demonstrates the effects of retrieving the relative position vector in the local frame of reference. $V^t$ and $V^{t+1}$ represents the relative position vectors at time $t$ and $t+1$, respectively in the global frame of the reference. Note that despite the same lateral movement of the car in the 2 examples, these vectors have different coordinates. However, after projecting them in their respective local frames of reference, the new vectors $dx$ and $dy$ have the same values in the 2 cases.}
  \label{fig:relVector_in_refFrame}
\end{figure}

Note that the relative translation vector for a previous image gives the direction of motion of the car for the current time step. $V^t$ and $V^{t+1}$ are the relative translation vectors for the previous and current image in the global frame of the reference, respectively. We have already defined the local frame of reference such that the direction of the current motion of the car is in the $x'$-direction. Hence, in this new local frame, $V^t$ will be aligned with the $x'$-axis. The rotation matrix which aligns $V^t$ in the $x'$-direction can be formulated as:
\begin{align}
\frac {1}{||V^t||} \cdot
\begin{bmatrix}
\cos(\theta) & -\sin(\theta)  \\
\sin(\theta) & \cos(\theta)  
\end{bmatrix}
\begin{bmatrix}
V^t_{x} \\ V^t_{y}  
\end{bmatrix}
=
\begin{bmatrix}
1 \\ 0 
\end{bmatrix}
\end{align}

After having solved for the rotation matrix, we can multiply the vector $V^{t+1}$, to get $dx$ and $dy$ indicating the forward and lateral movements in the local frame of reference. 
\begin{align}
\begin{bmatrix}
\cos(\theta) & -\sin(\theta)  \\
\sin(\theta) & \cos(\theta)  
\end{bmatrix}
\begin{bmatrix}
V^{t+1}_x \\ V^{t+1}_y  
\end{bmatrix}
=
\begin{bmatrix}
dx \\ dy 
\end{bmatrix}
\end{align}
Figure~\ref{fig:relVector_in_refFrame} shows the effect of this transformation. Here, 2 examples with different vectors $V^t$ and $V^{t+1}$ (corresponding to the same steering command) in the global frame of reference have the same $dx$ and $dy$ in their respective local frame of reference. Alternatively, the angle $\theta$ can be determined from the cosine similarity between the vectors $V^t$ and $V^{t+1}$. The sign of $dy$ can be found by determining which side the vector $V^{t+1}$ is with respect to the line formed by the vector $V^t$. We train a neural network to predict the lateral movement $dy$, by minimizing the $L1$ loss between the prediction and this calculated label.

\section{Data Collection and Testing}
The CARLA simulator~\cite{DosovitskiyCoRL2017} data used for training was collected by running the car in autopilot mode at a frame rate of 30fps. The car is controlled by adjusting the throttle and steering command. The throttle influences the speed whereas the steering command controls the steering angle of the car. The throttle ranges between a value of 0 and 1. The car is at rest when the throttle is at zero and moves faster as it is increased to a maximum value of 1. In the autopilot mode, the mean throttle value is around 0.5. Moreover, the average speed at which the car executes the turn is around 20 km/h. This is within the range of values at which the car does not slip and therefore also matches with our assumption described in Section 2.2 of the main paper. The steering command varies between -1 and -1 with 1 corresponding to 70$^{\circ}$.

\noindent{\bf Visual Odometry:}

\begin{figure}
  \centering
  \includegraphics[width=0.5\textwidth]{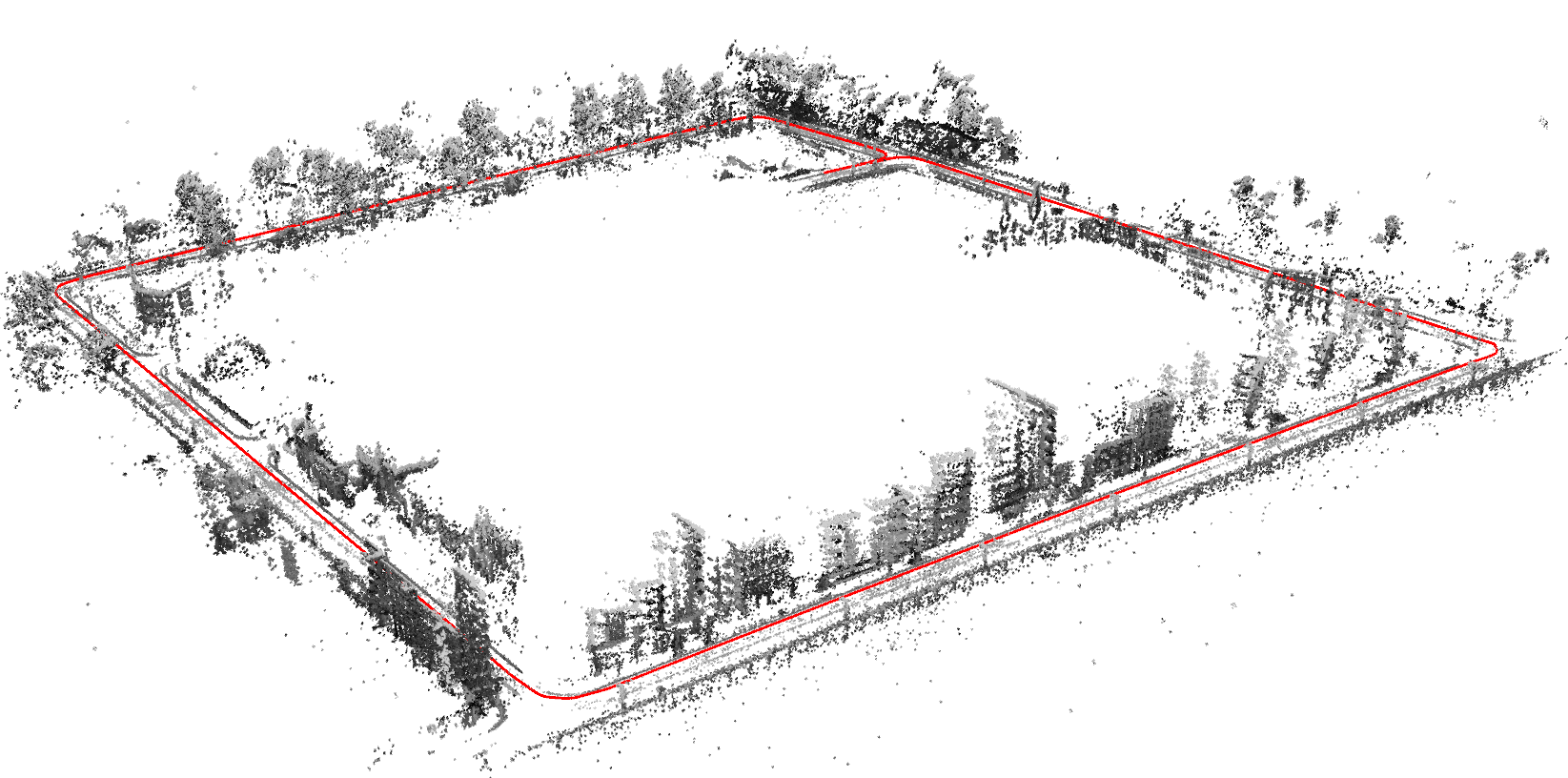}
  \caption{This figure shows a point cloud generated with visual odometry. We use the estimated poses (in red) to train our models.}
  \label{fig:pointcloud_side_view}
\end{figure}

During the data collection phase images of size $512 \times 512$ pixels are recorded along with the corresponding ground truth steering angles. Note that this ground truth steering angle data is only used to train the Oracle (supervised model). Whereas our model is trained with the visual odometry camera poses. Figure~\ref{fig:pointcloud_side_view} shows the trajectory of the estimated poses (in red) and the resulting point cloud generated when running Stereo DSO~\cite{wang2017stereoDSO}.

\noindent{\bf Front Wheel Steering:}

\begin{figure}[t]
  \centering
  \includegraphics[width=\linewidth]{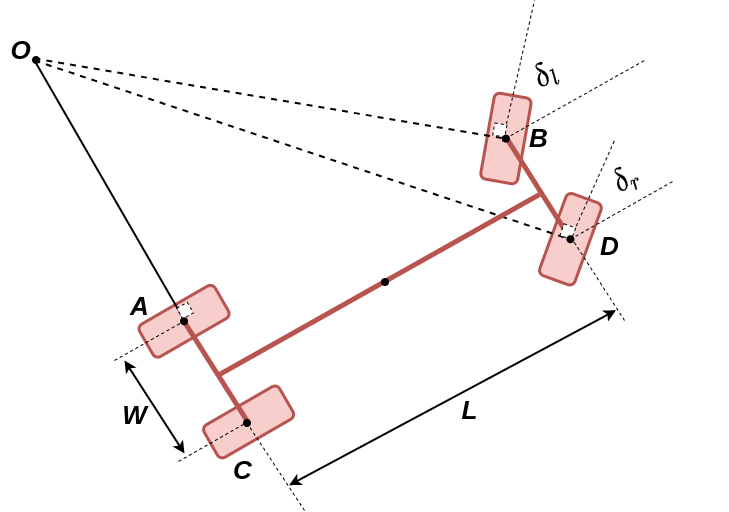}
  \caption{Depicts the steering angles of the 2 front wheels which are different as the car executes a turning maneuver. The 2 front wheels are oriented in a manner such that they have the same instantaneous center of rotation.}
  \label{fig:actual_car_model}
\end{figure}

\begin{figure}[t]
  \centering 
  \includegraphics[width=\linewidth]{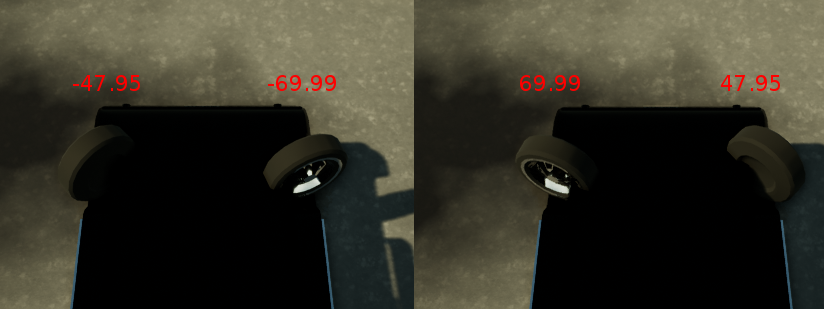}
  \caption{Depicts the orientation of the 2 front wheels when the wheel towards the inner circle of the radius of the turn is at the maximum value of approximately 70$^{\circ}$. Note that the outer wheel has a lower turning angle since it has to cover a larger curvature distance. Figure is taken from~\cite{gitissue}.}
  \label{fig:carla_front_steering}
\end{figure}

Note that in the bicycle model described in Section 2.2 the left and right front wheels were both modeled by a single front wheel with a steering angle $\delta$. As depicted in Figure \ref{fig:actual_car_model}, while executing a turning maneuver the left and right front wheels, will have slightly different steering angles denoted by $\delta_{l}$ and $\delta_{r}$ for the same instantaneous center of rotation $O$. When the car is making a left turn, than $\delta_{l} > \delta_{r}$ and vice versa, i.e., the steering angle of the inner front tyre would be greater than that of the outer front tyre. The difference can be approximated to be \cite{rajesh2012}:
\begin{align}
    \Delta\delta = \delta^2 \frac{W}{L}
\end{align}
Where, $L$ is the length of the wheelbase, and $W$ is the track width of the car and  
\begin{align}
    \delta =  \frac{\delta_{l} + \delta_{r} }{2}.
\end{align}
The Ackermann steering mechanism \cite{ackermann} can be used to ensure that the $\Delta\delta$ between the two front tyres is maintained while the car is making a turn. The CARLA simulator already caters for this difference and no additional correction needs to be performed. Figure \ref{fig:carla_front_steering} depicts the difference in steering angles of the front tyres when the inner tyre is at a maximum of 70$^{\circ}$

\noindent{\bf Model Architecture:}

\begin{figure}
  \centering
  \includegraphics[width=0.75\linewidth]{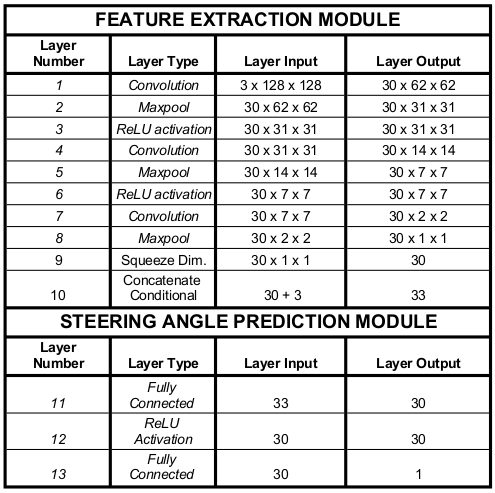}
  \caption{The convolution layers numbered 1, 4, and 7 have a kernel size of 5, with stride 2, and no additional padding. The number of kernels in each of these convolution layers is 30. The max pooling layers numbered 2, 5, and 8 have a kernel size of 2, and stride of 2 with no padding. The concatenation of the 3-dimensional vector is a one-hot encoding, indicating the car to turn left, right or keep move straight.}
  \label{fig:architecture}
\end{figure}

Note that this method uses stereo visual odometry, thereby also giving the notion of scale. However, the neural network model only requires a single image to predict the lateral component of the translation vector for the next frame at a fixed distance of $dx$ apart. The architecture of the network is described in Figure \ref{fig:architecture}. The image is downscaled to a lower resolution of $128 \times 128$, thereby simplifying the architecture. Note that the model comprises of a Feature Extraction Module (FEM) and a Steering Angle Prediction (SAP) Module. The FEM is a series of Convolution, Maxpooling and ReLU activation layers. Meanwhile, the SAP has 2 fully connected layers with one ReLU activation in between. The training was done with an initial learning rate of $0.0001$ using the Adam optimizer \cite{DBLP:journals/corr/KingmaB14}.

\noindent{\bf Testing at higher speeds:} 

\begin{figure}
  \centering
  \includegraphics[width=\linewidth]{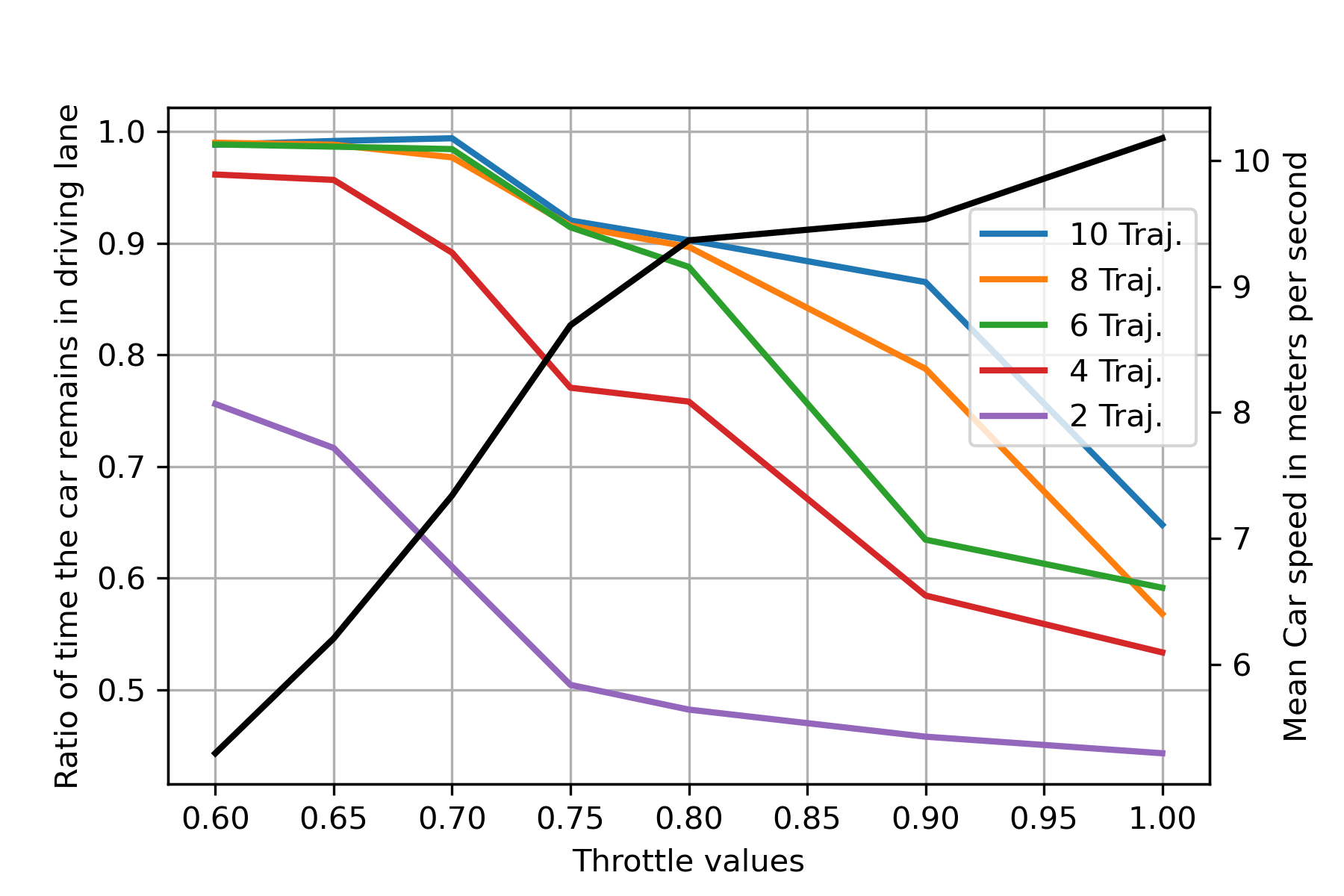}
  \caption{Shows the effect of gradually increasing the throttle on the mean speed of the car and the online performance for the models trained with 2, 4, 6, 8, and 10 trajectories. The speed of the car (right vertical axis) is reported in meters per second. The online performance (left vertical axis) is reported as the ratio of time the car remains within its driving track.}
  \label{fig:speed_increase}
\end{figure}

The equations derived in Section 2.2 of the main paper were based around the critical assumption that the car is moving at moderately low speeds. A car turning at 5 m/s will be subjected to very low lateral forces on the tyres and hence experience negligible slipping~\cite{rajesh2012}. While this assumption is reasonable in many urban scenarios, it may not hold in other circumstances. Therefore, we would like to assess how our models will behave if this assumption does not hold when turning. We successively enhance the throttle of the car, which leads to an increase in speed. The online performance of the models trained with 2, 4, 6, 8, and 10 trajectories is reported. This is depicted in Figure~\ref{fig:speed_increase}.

It can be observed that as the throttle is increased, so is the mean speed. Models trained with more trajectories are more robust to the speed than the ones trained with less number of trajectories. At a throttle value of 0.7, when the mean speed of the car is 7.5 m/s the models trained with 6 or more trajectories still maintain the same performance. Moreover, at a mean speed of around 9 m/s the performance of these high trajectory models only drops by about 10\%. This is despite an increase in speed by approx. 80\% from the assumption of 5 m/s. On the other hand, models trained with fewer trajectories show a dramatic drop in performance as the speed is successively increased. This demonstrates that training with more trajectories tends to be more robust in performance as the speed deviates farther away from our assumption. Nevertheless, an increase in the mean speed beyond 9 m/s leads to a significant drop in performance even for the models trained with a greater number of trajectories. To cater for this limitation, the lateral dynamics of the car would also need to be incorporated into the model to enable the car to perform stable high-speed turning maneuvers. While this is beyond the scope of this paper, we leave it for further work.

\noindent{\bf Testing on a new Town:}

\begin{figure}[ht]
  \centering
  \includegraphics[width=\linewidth]{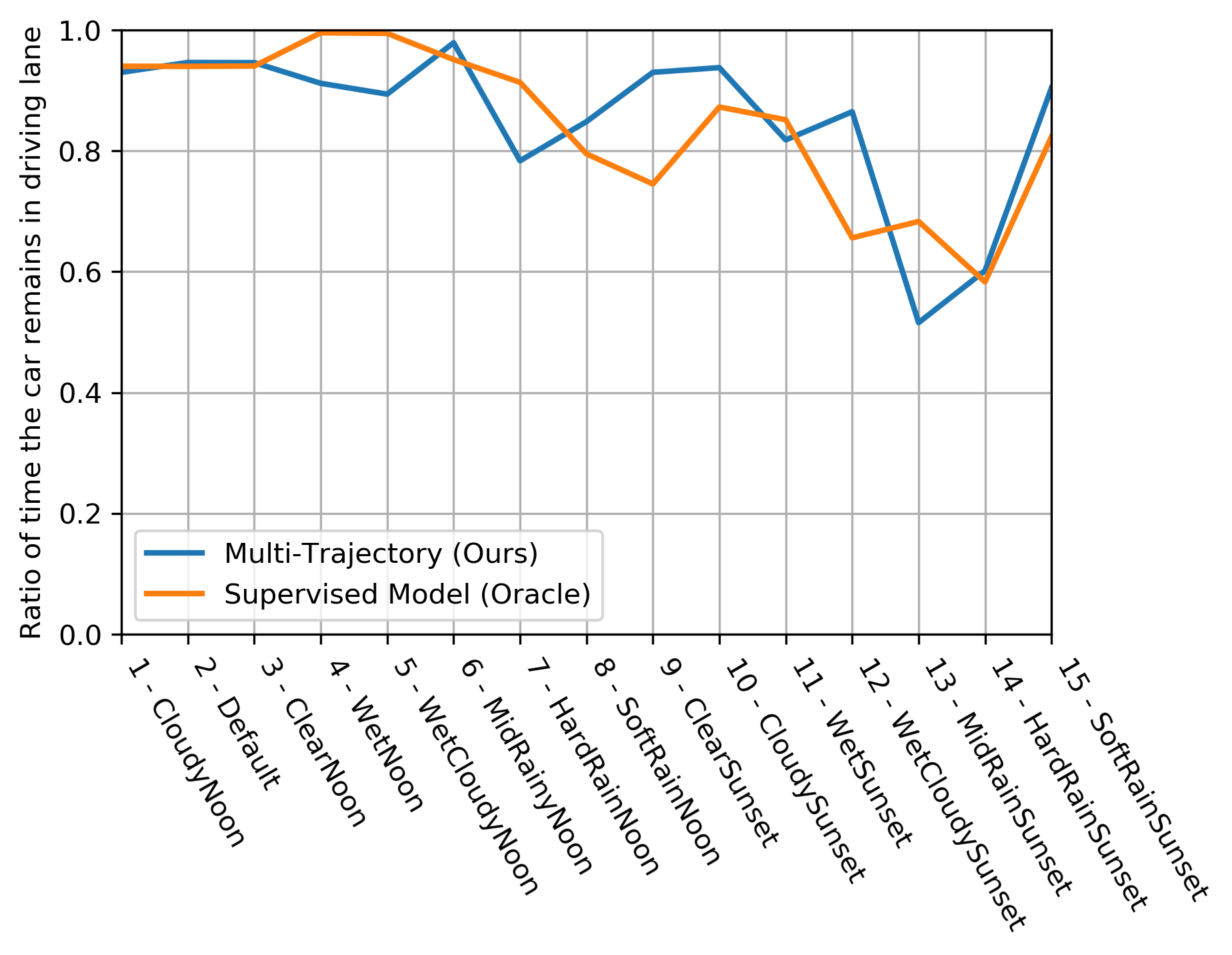}
  \caption{The plot exhibits the ratio of time the car remains within its own driving track for 2 models approach across all the 15 different weather conditions. Both models were trained on Town2 but evaluated on Town1. Higher is better.}
  \label{fig:different_towns}
\end{figure}

The results from the main paper show that our self-supervised framework for steering angle prediction is comparable to the supervised method. The models were trained and evaluated in the same Town albeit across different weathers. However, generalization to unseen environments is also important. CARLA v0.8.2 provides 2 different towns. Therefore, in this supplementary material, an additional experiment is performed. Our method and the supervised model are trained on Town2 but evaluated on Town1. Figure \ref{fig:different_towns} shows the performance of both methods across all 15 weather conditions. The performance of both methods in an unseen environment drops slightly in comparison to when trained and evaluated in the same towns. However, it is important to observe that our method is still on par with the supervised method. This aligns with the results from the main paper.

\section{Video}

The video at \url{https://youtu.be/_22gF9GuwTg} shows the performance between the visual odometry 
models trained with one and trained with multiple trajectories. As can be observed, the multiple trajectory 
model is capable of recovering the course despite some deviations from the reference. This is in contrast 
to the model trained with only one VO trajectory.

\bibliography{supplement}